\documentclass[10pt,twocolumn,letterpaper]{article}

\usepackage[pagenumbers]{cvpr} % To force page numbers, e.g. for an arXiv version

\usepackage{graphicx}
\usepackage{amsmath}
\usepackage{amssymb}
\usepackage{booktabs}
\usepackage{multirow}
\usepackage{pifont}

\usepackage[pagebackref,breaklinks,colorlinks]{hyperref}

\usepackage[capitalize]{cleveref}
\crefname{section}{Sec.}{Secs.}
\Crefname{section}{Section}{Sections}
\Crefname{table}{Table}{Tables}
\crefname{table}{Tab.}{Tabs.}

\def\cvprPaperID{*****} % *** Enter the CVPR Paper ID here
\def\confName{CVPR}
\def\confYear{2023}

\begin{document}

%%%%%%%%% TITLE - PLEASE UPDATE
\title{SynthASpoof: Developing Face Presentation Attack Detection Based on Privacy-friendly Synthetic Data}

\author{Meiling Fang$^{1,2}$, Marco Huber$^{1,2}$, Naser Damer$^{1,2}$ \\
$^{1}$Fraunhofer Institute for Computer Graphics Research IGD,
Darmstadt, Germany\\
$^{2}$Department of Computer Science, TU Darmstadt,
Darmstadt, Germany\\
Email: {meiling.fang@igd.fraunhofer.de}
}

\maketitle

%%%%%%%%% ABSTRACT
\begin{abstract} % ready for check
Recently, significant progress has been made in face presentation attack detection (PAD), which aims to secure face recognition systems against presentation attacks, owing to the availability of several face PAD datasets. However, all available datasets are based on privacy and legally-sensitive authentic biometric data with a limited number of subjects. 
To target these legal and technical challenges, this work presents the first synthetic-based face PAD dataset, named SynthASpoof, as a large-scale PAD development dataset. 
The bona fide samples in SynthASpoof are synthetically generated and the attack samples are collected by presenting such synthetic data to capture systems in a real attack scenario. 
The experimental results demonstrate the feasibility of using SynthASpoof for the development of face PAD.
Moreover, we boost the performance of such a solution by incorporating the domain generalization tool MixStyle into the PAD solutions. 
Additionally, we showed the viability of using synthetic data as a supplement to enrich the diversity of limited authentic training data and consistently enhance PAD performances. The SynthASpoof dataset, containing 25,000 bona fide and 78,800 attack samples, the implementation, and the pre-trained weights are made publicly available \footnote{\url{https://github.com/meilfang/SynthASpoof.git}}.

\end{abstract}

% dataset name 
% facial attributes

%%%%%%%%% BODY TEXT
\section{Introduction} % ready for check - MH Done
\label{sec:intro}

% what is PAD, why PAD is important, current algorithms -> the problem, (data) -> existing dataset drawback -> privacy situtaion in face reognition models -> PAD attack synthetic (difference with deepface) -> our contribution

Due to its outstanding performance, face recognition has been widely used in various aspects of our daily lives, such as access control, phone unlocking, and mobile payments. However, face recognition is vulnerable to presentation attacks (PAs) including print attacks, video replay attacks, and 3D mask attacks \cite{casia_fas,replay_attack,oulu_npu,padisi-Face}. Therefore, face presentation attack detection (PAD), referring to the process of identifying whether a face presented to the system is a bona fide (live) or PA (spoof), is essential to secure face recognition from PAs. 

With the advancements in deep learning technology, face PAD algorithms have made great progress. One of the main contributors to this advance is the face PAD datasets \cite{casia_fas,oulu_npu,replay_attack,DBLP:conf/eccv/ZhangYLYYSL20,DBLP:conf/cvpr/LiuSJ019}.
However, these datasets utilized for developing data-driven PAD solutions are built on authentic biometric data, which might raise ethical and legal challenges.
This concern has recently been discussed in both the face recognition \cite{DBLP:conf/iccv/QiuYG00T21} and face morphing attack detection \cite{DBLP:conf/cvpr/DamerLFSPB22} communities. Given the legal privacy regulations, the collection, use, share, and maintenance of face data for biometric processing is extremely challenging \cite{onsyntheticdata_20220922}. For example, several large-scale face recognition datasets, such as VGGFace2 \cite{DBLP:conf/fgr/CaoSXPZ18}, MS-Celeb-1M \cite{DBLP:conf/eccv/GuoZHHG16}, and MegaFace \cite{DBLP:conf/cvpr/Kemelmacher-Shlizerman16}, were withdrawn by their creators with privacy and proper subjects consent issues being the main drive. 
One of the main candidate solutions for this issue is the use of synthetic data \cite{onsyntheticdata_20220922}. This has been very recently and successfully proposed for the training of face recognition \cite{DBLP:conf/iccv/QiuYG00T21,DBLP:conf/icb/BoutrosHSRD22,DBLP:conf/fgr/BoutrosKFKD23} and morphing attack detection \cite{DBLP:conf/cvpr/DamerLFSPB22,DBLP:conf/icb/HuberBLRRDNGSCT22,DBLP:conf/icb/FangBD22}, among other processes such as model quantization \cite{DBLP:conf/icpr/BoutrosDK22}. 
%issue has been successfully addressed in several biometric use cases, such as face image quality \cite{DBLP:conf/iwbf/ZhangGRR021}, face recognition \cite{DBLP:conf/iccv/QiuYG00T21,DBLP:conf/icb/BoutrosHSRD22}, and face morphing attack detection \cite{DBLP:conf/cvpr/DamerLFSPB22,DBLP:conf/icb/HuberBLRRDNGSCT22}, by using synthetic data.
%Zhang \etal \cite{DBLP:conf/iwbf/ZhangGRR021} studied the behavior of face image quality assessment methods on synthetic images generated by StyleGAN and compared the face image quality values with those of authentic face images.
%Qiu \etal \cite{DBLP:conf/iccv/QiuYG00T21} explored the performance gap between face recognition models trained on authentic and synthetic data and proposed an identity mixup and domain mixup method to mitigate the gap.
%Damer \etal \cite{DBLP:conf/cvpr/DamerLFSPB22} introduced a synthetic-based morphing attack detection (MAD) development dataset (SMDD) for training MAD models and the results demonstrated the applicability of the SMDD for developing generalized MAD.
%These works inspired us to build a synthetic-based face PAD dataset for developing PADs. 
Synthetic data for PAD development has, besides the privacy and legal motivations, a major advantage when it comes to scale and diversity.
While most existing face PAD datasets are of a small-scale with a limited number of subjects, creating a synthetic-based PAD development data enables to produce a large-scale dataset in terms of both, the number of samples and the number of different faces.

%The other motivation of synthetic PAD data is that most existing face PAD datasets are small-scale with a limited number of subjects and thus do not fulfill the training needs for over-parameterized deep learning models.

Motivated by the legal and ethical challenges in using, sharing, and collecting authentic biometric data along with the limitation in the scale and diversity in existing datasets, this work poses the question of "can synthetic data be used for the development of face PAD?".
This is based on our assumption that learning to detect the differences between bona fide and attack samples of a synthetic origin can be used to detect these differences between authentic bona fide and attacks, and thus perform PAD. 
Towards that, we introduce \textbf{the first privacy-friendly synthetic-based face PAD (Anti-Spoofing) dataset, SynthASpoof}, consisting of 25,000 bona fide and 78,800 attack samples. 
The bona fide samples are created by using StyleGAN2-ADA \cite{DBLP:conf/nips/KarrasAHLLA20}, while the attack samples are collected by presenting these synthetic samples as printed/replayed attacks to varied capture sensors.
%To overcome the problems posed by the need for diverse training datasets and the legal considerations of using, reusing, and sharing biometric data, we introduce \textbf{the first privacy-friendly synthetic-based face PAD (Anti-Spoofing) dataset, SynthASpoof}, consisting of 25,000 bona fide and 78,800 attack samples. 
%The bona fide samples are \cite{DBLP:conf/cvpr/DamerLFSPB22} created by using StyleGAN2-ADA \cite{DBLP:conf/nips/KarrasAHLLA20}, while the attack samples are collected by presenting the printed/replayed synthetic images to varied capture sensors.
% preserve the discriminative intrinsic attack clues (e.g., material, reflection, and geometric distribution).
Samples of the SynthASpoof are shown in Fig.  \ref{fig:synPADsamples}. 
Based on the SynthASpoof dataset, we then conduct extensive experiments to \textbf{explore the feasibility of using synthetic data for the development of face PADs}. Subsequently, we propose to \textbf{adapt MixStyle \cite{DBLP:conf/iclr/ZhouY0X21} to enhance the generalizability of models trained on SynthASpoof}. Furthermore, we \textbf{successfully propose supplementing the authentic training data with the synthetic SynthASpoof to achieve even higher PAD performances}. %\textbf{investigation on a combination of authentic and synthetic PAD datasets} to demonstrate the possibility of the SynthASpoof dataset for face PADs.

\section{Related Work} % ready for check - MH Done
\label{sec:related_work}

%=====================================================================
\begin{table*}[htb]
\centering
\resizebox{0.9\textwidth}{!}{
\begin{tabular}{c|c|c|c|c}
\hline 
Dataset & Year & \# Bona fide/attack  & \# Sub & Attack types  \\ \hline \hline
CASIA-FASD \cite{casia_fas} & 2012 & 150 / 450 (V) & 50 & 1 Print, 1 Replay \\ 
Replay-Attack \cite{replay_attack} & 2012 & 200 / 1,000 (V) & 50  & 1 Print, 2 Replay   \\ 
MSU-MFSD \cite{msu_mfs} & 2015 & 70 / 210 (V) & 35 & 1 Print, 2 Replay \\
%HKBU-MARs \cite{hkbu} & 2016 & 120 / 60 (V) & 12  & 2 3D masks  \\ \hline
OULU-NPU \cite{oulu_npu}  & 2017 & 1,980 / 3,960 (V) & 55 & 2 Print,2 Replay \\ 
SiW-M \cite{DBLP:conf/cvpr/LiuSJ019} & 2019 & 660 / 968 (V) & 493$^*$ & 1 Print, 1 Replay, 5 3D Mask, 3 Make Up, 3 Partial \\ 
CelebA-Spoof \cite{DBLP:conf/eccv/ZhangYLYYSL20} & 2020 & 184,407 / 377,168(I)       & 10,177$^*$  & 3 Print, 3 Replay, 1 3D, 3 Paper Cut \\ 
PADISI-Face \cite{padisi-Face}  & 2021 & 1,105 / 924 (V) & 360 & 1 Print, 4 Mask, 1 Makeup, 1 Tattoo, 2 Partial \\ \hline
SynthASpoof & 2023 & 25,000 / 78,800 (I\&V)  & 25,000 & 1 Print, 3 Replay \\ \hline
\end{tabular}}
\caption{Summary of the public face PAD datasets. V and I are shorthand for video and image, respectively. Subject number with '*' denotes the subjects are partially or all from the web. Note the limited scale of most datasets and the fact that the larger ones are based on web-collected images.}
\label{tab:datasets-summariz}
\end{table*}
%=====================================================================

In the last decade, many face PAD datasets \cite{padisi-Face,DBLP:conf/eccv/ZhangYLYYSL20,DBLP:conf/cvpr/LiuSJ019,oulu_npu,msu_mfs,replay_attack,casia_fas,DBLP:journals/tifs/LiuZYWSLTEXLGLL22} have been collected and made available to support the development of PAD algorithms. The face PAD datasets can be broadly categorized into four groups based on the type of attacks and sensors: multi-modal 3D attacks \cite{DBLP:journals/tifs/GeorgeMGNAM20}, multi-modal 2D attacks \cite{DBLP:conf/cvpr/ZhangWLZ0ESWL19,padisi-Face}, single-modal 3D attacks \cite{DBLP:journals/tifs/LiuZYWSLTEXLGLL22}, and single-modal 2D attacks \cite{oulu_npu,msu_mfs,replay_attack,casia_fas}.
%However, they were all based on limited and privacy-sensitive authentic data.

Multi-modal datasets \cite{DBLP:conf/cvpr/ZhangWLZ0ESWL19,DBLP:journals/tifs/GeorgeMGNAM20} used multiple sensors in addition to visible cameras, such as depth and infrared cameras, providing more options for face PAD solutions. However, such datasets have limitations in real-world deployment due to the cost of sensors and computation resources. 3D attacks are more realistic than traditional 2D attacks. The HiFiMask \cite{DBLP:journals/tifs/LiuZYWSLTEXLGLL22} dataset is the largest and most recent 3D face mask PAD dataset, collected from 75 subjects and including three mask attacks. However, HiFiMask dataset has a limited number of subjects and mask materials due to the higher cost of 3D mask creation compared to 2D attacks. Most 2D face PAD datasets (as seen in Table \ref{tab:datasets-summariz}) are outdated due to their acquisition equipment and have limited numbers of subjects and samples, leading to a potential over-fitting risk.
SiW-M \cite{DBLP:conf/cvpr/LiuSJ019}, PADISI-Face \cite{padisi-Face}, and CelebA-Spoof \cite{DBLP:conf/eccv/ZhangYLYYSL20} are relatively up-to-date and large-scale datasets, where CelebA-Spoof and part of the SiW-M dataset were collected from the web. Many of the existing face PAD datasets have ethical and legal issues that limit their public availability and raise concerns about sharing and reusing biometric information of individuals, driving some researches to keep their developed datasets private \cite{DBLP:journals/pr/FangDKK22,DBLP:conf/fgr/FangBKD21}. For example, SiW-M is currently inaccessible.  In addition to privacy issues, CelebA-spoof has several limitations: 1) numerous label noise, 2) low quality attack samples, which contradict the fact that attackers commonly use highly sophisticated artifacts to maximize their impersonation success probability, and 3) no consent of all involved individuals. 

Overall, existing face PAD datasets have two primary limitations. First, the collection, use, and share of such data pose ethical and legal challenges \cite{onsyntheticdata_20220922}. Second, the scale of the existing face PAD datasets may not be sufficient to develop over-parameterized deep learning based PAD solutions.
This highlights the need for face PAD development datasets that prioritize the privacy of individuals, the shareability of data in the research community, and the reproducibility and continuity of face PAD research. 
To address these concerns, we propose the use of synthetic data for the development of face PAD. Our synthetic SynthASpoof data contains of 25,000 bona fide samples and 78,800 attack samples (details in Section \ref{sec:SynthASpoof}).

\section{SynthASpoof Dataset} % ready for check - MH Done
\label{sec:SynthASpoof}

\begin{figure}[th!]
\begin{center}
\includegraphics[width=0.92\linewidth]{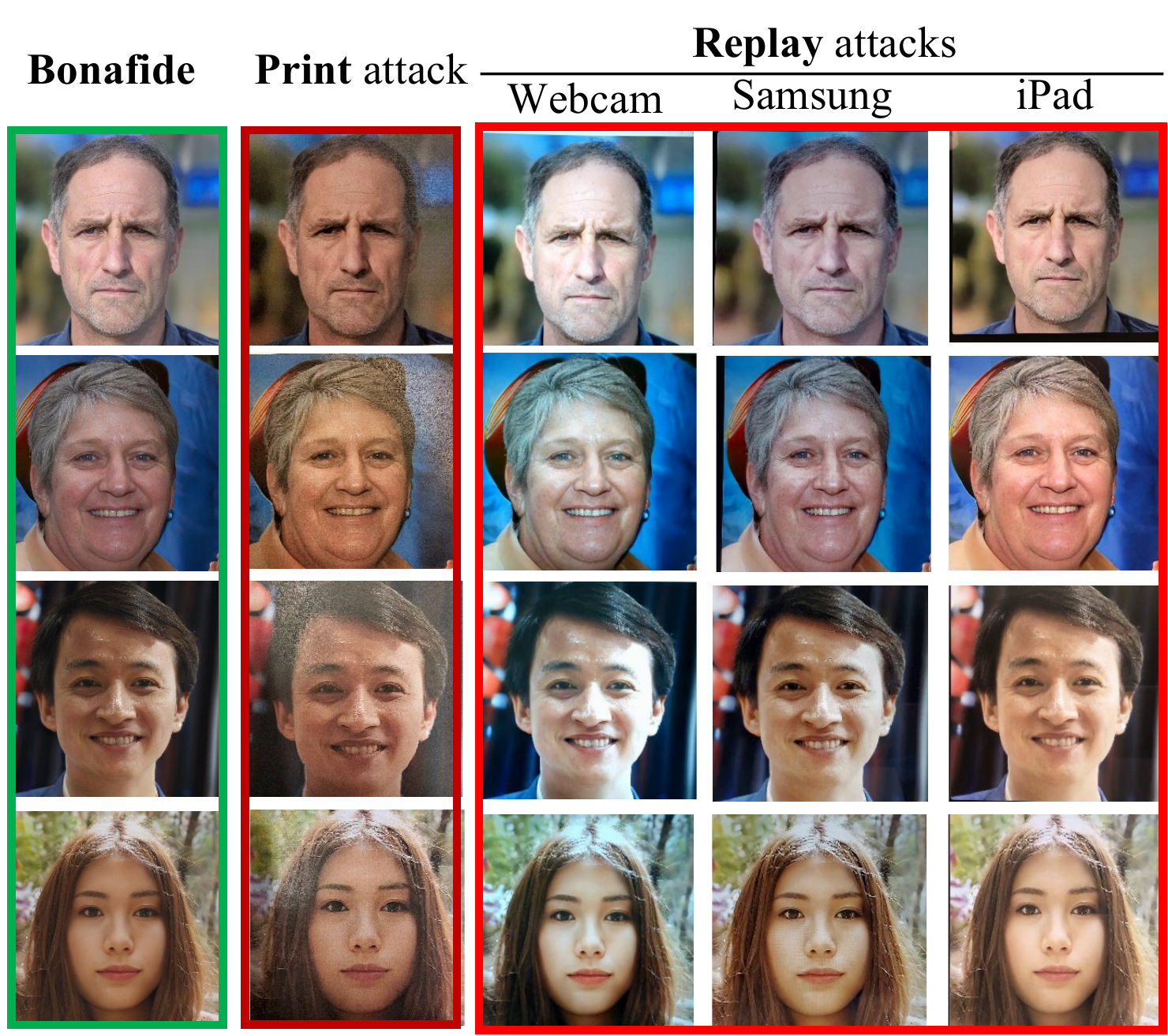}
\caption{Samples of the SynthASpoof dataset. The left column shows bona fide samples. The second to last column show different attack samples collected from the corresponding bona fide images. In the case of replay attacks, three sensors (webcam, Samsung phone, and iPad) were used to capture the attacks displayed on different screens. }
\label{fig:synPADsamples}
\vspace{-5mm}
\end{center}
\end{figure}

Despite the significance of publicly available datasets in promoting the progress of face PAD and being valuable sources for the research community, legal, ethical and privacy concerns, as well as the  limited size and diversity of the datasets pose challenges to the development of generalized PAD solutions.

This section introduces our SynthASpoof dataset (samples are shown in Fig.  \ref{fig:synPADsamples}), which is the first synthetic-based face PAD dataset. The dataset is built based on the image synthesis and selection procedure presented in \cite{DBLP:conf/cvpr/DamerLFSPB22}.
%, where all the bona fide samples in this datasets are from the training set of SMDD. 
%The dataset is created based on the SMDD dataset \cite{DBLP:conf/cvpr/DamerLFSPB22}, as all the bona fide in the SynthASpoof dataset are samples from the training set of the SMDD dataset.
To follow realistic attack scenarios appearing in authentic data attacks, the attack samples are created based on the synthetic bona fide data by presenting printed and replayed images to capture sensors. 
This aims at fulfilling our assumption that the difference between authentic bona fide and authentic-based attacks induced by the attack process can also be induced by the same attack process using synthetic data. Thus, learning to detect this difference on synthetic data will enable detecting it in authentic-based attacks. 
%To preserve the attack cues, such as moire pattern and color distortion, the attack samples are created based on the bona fide data by presenting printed and replayed images to capture sensors.

\textbf{Bona Fide Samples:} 
First, 125,000 images were created by using the StyleGAN2-ADA \cite{DBLP:conf/nips/KarrasAHLLA20} trained on Flickr-Faces-HQ dataset (FFHQ) \cite{DBLP:conf/cvpr/KarrasLA19}. The pretrained model produced a synthetic face data for each latent vector that was randomly non-repeatedly generated based on Gaussian noise. 
These images were then filtered automatically by using the CR-FIQA \cite{DBLP:journals/corr/abs-2112-06592} face image utility assessment approach, where extreme non-frontal poses and largely occluded images were mostly removed by removing the images with the lowest utility score. This helps simulate the real log-in face recognition scenario that is commonly targeted by PAs.
Finally, SynthASpoof contains 25,000 bona fide samples.
%The 25,000 synthetic bona fide samples were created by using the StyleGAN2-ADA \cite{DBLP:conf/nips/KarrasAHLLA20} trained on Flickr-Faces-HQ dataset (FFHQ) \cite{DBLP:conf/cvpr/KarrasLA19}. 
%These images were filtered automatically from 125K generated images from random non-repeated Gaussian noise by using the CR-FIQA \cite{DBLP:journals/corr/abs-2112-06592} face image utility assessment approach, where extreme non-frontal poses and largely occluded images were mostly removed by removing the images with the lowest utility score. This helps simulate the real log-in face recognition scenario that is commonly targeted by PAs. 

%The domain gap is highly correlated to the key factor of recognizing spoof: visual appearance. Spoofing cues, such as moire pattern and color distortion, can dramatically change or disappear with different camera devices, illuminations, and image resolutions. For example, images from Oulu-NPU [5] are in 1080P resolution, while images from Idiap Replay [9] are only in 480P resolution. The sensor noise and low image quality of Idiap Replay can lead to a biased prediction as spoof from a model trained on Oulu-NPU.

\textbf{Attack Samples:} SynthASpoof contains two attack types, print and replay attacks. 
For the print attacks 3,800 videos of distinct synthetic subjects were captured using a Samsung Galaxy Tablet S6.
%We obtained 3,800 videos of different synthetic subjects by using a Samsung Galaxy Tablet S6 to capture the printed photos. 
%Compared to labor-intensive print attack collection, replay attacks were collected using more diverse electronic equipment. 
For the more challenging replay attack, we introduce diverse display and capture setups.
First, attacks displayed on a MacBook Air 2020 screen were captured using both a Samsung Galaxy A71 and an iPad Pro 10.5 (both with a resolution of $1920 \times 1080$).
Additionally, attacks displayed on a Dell UltraSharp 24 display were captured using a Creative Labs webcam with a resolution of $720 \times 480$.
All the 25,000 images were used as an attack on each setup resulting in a total of 75,000 replay attack clips.
All attack captures (print and replay) were cropped so that they do not include any region outside of the displayed attack image (e.g. screen border). 
All captured attacks are videos with a duration ranging from 3 to 5 seconds, from each of these videos the single frame in the middle of the video is also extracted as a single image attack used in our training.
%A Samsung Galaxy A71 and iPad Pro 10.5 (both with a resolution of $1920 \times 1080$) were utilizied to capture replayed bona fide images from a MacBook Air 2020 screen, while a XXX webcam with a resolution of $720 \times 480$ was used to capture images from a Dell UltraSharp 24 display. 
%Therefore, each subject corresponded to 3 replay attack samples, i.e., a total of 75,000 replay attack clips. 
%All video clips were pre-processed to remove background information and were cut roughly in 3-5 seconds clips. 
%For each video, the middle frame was extracted for further study. 
%Both video and image format will be released to provide more possibilities to develop PAD solutions.
Both, the videos and the images of the SynthaSpoof dataset are publicly released and can be used to develop PAD solutions based on synthetic data.
% (Link redacted for blind review)

Comparing to existing face PAD datasets, the proposed SynthASpoof dataset provides three advantages: \textbf{1) Privacy-friendly:}  SynthASpoof is the first synthetic face PAD dataset which relaxes the pure dependence on the legally and ethically challenging use of authentic development data. \textbf{2) Large-scale and high-quality samples}: 
As discussed in Section \ref{sec:related_work}, most existing datasets are of small scale, the only relatively larger and diverse dataset is the CelebA-Spoof \cite{DBLP:conf/eccv/ZhangYLYYSL20}. However, some of its bona fide samples might be falsely annotated and many of the attack samples exhibit severe distortion and low quality, as they were collected from the web (which is a orivacy issue by itself) without proper control or post-processing checks. 
In contrast, the bona fide samples in SynthASpoof were checked by face image quality control and the attack samples were collected in a controlled manner to reflect the fact that attackers usually use highly sophisticated artifacts to maximize their success in impersonation. \textbf{3) Extensibility:} researchers can build subsequent synthetic-based face PAD datasets by increasing the diversity of attack types. % based on the provided bona fide synthetic images.

\section{PAD Solutions} % ready for check - MH Done
\label{sec:pad_solutions}
To assess the suitability of using SynthASpoof for the development of face PAD, we adopt two of the commonly used face PAD backbones, ResNet \cite{DBLP:conf/cvpr/HeZRS16} and PixBis  \cite{DBLP:conf/icb/GeorgeM19}. The selection of these two backbones was based on their wide use, representing two common supervision strategies in face PAD, and reported good performance in previous studies \cite{DBLP:conf/cvpr/HeZRS16,DBLP:conf/icb/GeorgeM19,DBLP:journals/tbbis/YuLSXZ21,Fang_2022_WACV}. 

\subsection{Base Presentation Attack Detectors}
\textbf{ResNet} \cite{DBLP:conf/cvpr/HeZRS16} is one of the most popular backbone architectures used in face PAD algorithm design \cite{DBLP:journals/tbbis/YuLSXZ21,Fang_2022_WACV,DBLP:conf/eccv/ZhangYLYYSL20,DBLP:conf/icb/FangAKD22}. We report the results of a model trained from scratch based on the ResNet-18 model architecture. A cross-entropy loss function is used in the training phase and formulated as follows:
\begin{equation}
    \mathcal{L}_{CE} = -[y \cdot \log p + (1-y)\cdot \log (1-p)] ,
\end{equation}
where $y$ is the ground truth (1 for bona fide and 0 for attack in our case) and $p$ is the predicted score.

\textbf{PixBis} \cite{DBLP:conf/icb/GeorgeM19} employs a binary supervisory strategy at pixel-level to simplify the problem and obviate the need for a computationally intensive synthesis of depth maps. Two dense blocks of DenseNet121 \cite{DBLP:conf/cvpr/HuangLMW17} are utilized as the model backbone and a combination of two binary cross-entropy loss functions is used to train the model for both pixel-wise and binary output. 
The combined loss equation for the training of all models is formed as:
\begin{equation}
    \mathcal{L}_{PixBis} = \mathcal{L}_{CE}^{pixel-wise} + \mathcal{L}_{CE}^{binary}.
\end{equation}
where $\mathcal{L}_{CE}^{pixel-wise}$ refers to the loss based on the pixel-wise output and $\mathcal{L}_{CE}^{binary}$ refers to the loss based on the binary output.

\subsection{MixStyle}
Recent studies in synthetic-based face recognition revealed a domain gap between synthetic and authentic face images through the examination of the performance divergence between face recognition models trained on synthetic and authentic data. This performance gap has also been observed in our work, as will be detailed in Section \ref{ssec:performance_gap}. To narrow the domain gap between synthetic and authentic face PAD data, we adapt a recently proposed domain generalization method, MixStyle \cite{DBLP:conf/iclr/ZhouY0X21}. MixStyle mixes the feature statistics of two samples to synthesize novel domains inspired by the observation that the feature statistics encode style/domain-related information. To adapt the face PAD model from synthetic data to authentic data, we utilized the labeled synthetic SynthASpoof and unlabeled authentic face PAD data to perform MixStyle within a mini-batch and with a controlled probability during the training process. 
%Therefore, a mini-batch contains both the original synthetic and real stylized synthetic PAD data.

Mathematically, the MixStyle adapted in our case can be formulated as follows:
\begin{equation}
\begin{aligned}
    \gamma = \lambda \sigma(x_{s}) + (1-\lambda) \sigma(x_{a}) 
    \\
    \beta = \lambda \mu(x_{s}) + (1-\lambda) \mu(x_{a}) 
\end{aligned}   
\end{equation}
Where $x_s$ and $x_a$ refer to synthetic and authentic face PAD data, respectively. $\lambda \in \mathbb{R}^B$ are weights sampled from the Beta distribution.
The final mixed feature statistic is applied to the styled normalized synthetic face PAD data $x_s$ as:
\begin{equation}
    MixStyle(x_s) = \gamma  \frac{x_s-\mu(x_s)}{\sigma(x_s)} + \beta
\end{equation}
% Question: Using originial defined MixStyle or StylizedSyn or 
It is important to note that: 1) The loss is calculated only on the synthetic face PAD data output when conducting the domain adaptation experiment in Section \ref{ssec:effect_mixstyle}. 
%2) In Section \ref{ssec:limited_data}, where the model is trained on a combined dataset of labeled SynthASpoof and a real face PAD dataset and tested on an unseen real dataset to address the limited data problem, MixStyle is adapted to reduce the distance between the trained synthetic data and the real data. Specifically, the loss is calculated on both training datasets. 
2) In Section \ref{ssec:limited_data}, the face PAD model is trained on a combination of the SynthASpoof and an authentic face PAD dataset to address the problem of limited training data. MixStyle is used there to reduce the difference between the synthetic and authentic training data, which is demonstrated later by the performance on unseen authentic data. %authentic
3) MixStyle is removed during the inference process, and thus does not require additional computational overhead while using the PAD. 

In our experiment, MixStyle is inserted after the first and second ResNet blocks and after the first dense block of the PixBis model, based on the fact that features at higher layers have a stronger correlation with class labels as opposed to the domain information \cite{DBLP:conf/iclr/ZhouY0X21}.
% 3) MixStyle is removed during the inference process. To be more precise, in our experiments MixStyle is inserted after the first and second ResNet blocks and after the first dense block of the PixBis model, based on the fact that features at higher layers have a stronger correlation with class labels as opposed to the domain information.

\section{Experiments} % ready for check - MH Done
\label{sec:experiment}

\subsection{Datasets}

\begin{figure}[th!]
\begin{center}
\includegraphics[width=0.99\linewidth]{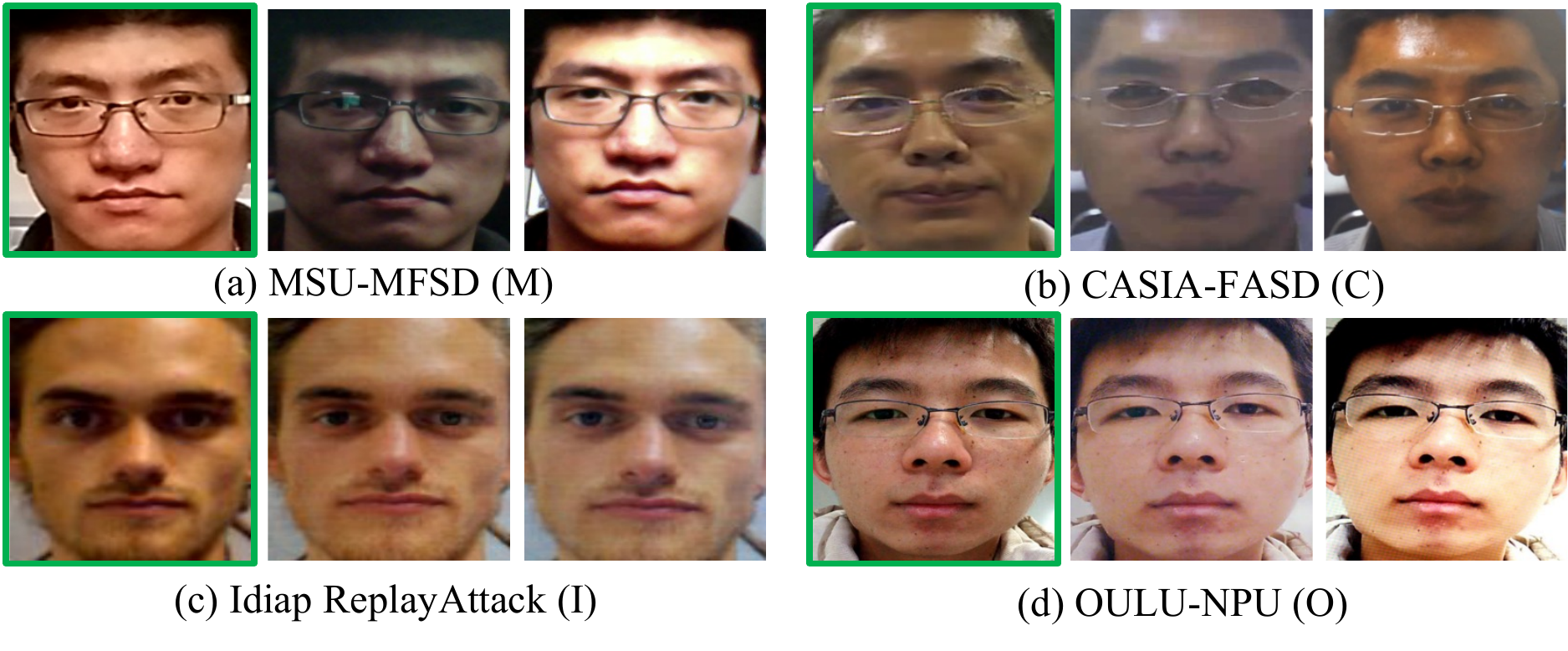}
\caption{Samples of four authentic face PAD datasets. Images with green bounding box are bona fide, while others are attack samples.}
\label{fig:real_pad_data}
\vspace{-5mm}
\end{center}
\end{figure}

To assess the feasibility of using SynthASpoof to develop face PAD, the performance of models trained on SynthASpoofis evaluated on four authentic face PAD benchmarks: MSU-MFSD \cite{msu_mfs} (denoted as M), CASIA-MFSD \cite{casia_fas} (denoted as C), Idiap Replay-Attack \cite{replay_attack} (denoted as I), and OULU-NPU \cite{oulu_npu} (denoted as O). The data samples are shown in Fig.  \ref{fig:real_pad_data}.
%We selected these datasets because they include a wide range of print and replay attacks.

The \textbf{MSU-MFSD} \cite{msu_mfs} dataset consists of 440 videos captured from 35 subjects using two different resolutions of cameras. The dataset contains two types of attacks, printed photo attacks and replay attacks.
The \textbf{CASIA-MFSD} \cite{casia_fas} dataset is comprised of 600 videos from 50 subjects and includes three types of attacks: warped photo attack, cut photo attack, and video replay attack.
The \textbf{Idiap Replay-Attack} dataset \cite{replay_attack} contains 300 videos from 50 subjects captured under various sensors and illumination conditions. The dataset includes two attack types: print attacks and replay attacks.
The \textbf{Oulu-NPU} \cite{oulu_npu} is a mobile face PAD dataset designed for assessing the generalizability of PAD methods in a realistic mobile scenario. OULU-NPU consists of 5940 video clips from 55 subjects using six different mobile phones. 

The performance of models trained on synthetic and authentic face PAD data is analyzed over these datasets. As SynthASpoof database is specifically created for the purpose of training face PAD models, the entire dataset is used in the training phase. The trained models are then further tested on other authentic face PAD datasets. 

\subsection{Implementation Setup}
Following \cite{Fang_2022_WACV,DBLP:conf/icb/GeorgeM19,DBLP:journals/tbbis/YuLSXZ21} and to make up for the relatively smaller size of the authentic datasets, 25 frames (per video) were sampled evenly across the duration of each video in the four authentic face PAD datasets, while only one frame was considered from each video in the SynthASpoof database as detailed in Sec. \ref{sec:SynthASpoof} and Tab. \ref{tab:datasets-summariz}. 
The faces were then detected and cropped using the MTCNN method \cite{DBLP:journals/spl/ZhangZLQ16} and resized to $224 \times 224 \times 3$ pixels, following \cite{DBLP:journals/tbbis/YuLSXZ21,Fang_2022_WACV,DBLP:conf/eccv/ZhangYLYYSL20,DBLP:conf/icb/GeorgeM19,DBLP:journals/tifs/WangHSC21}. 
During training, a weighted sampling was performed to insure a bona fide-attack ratio of 1:1.
The Stochastic Gradient Descent (SGD) optimizer with a momentum of 0.9 and weight decay of 5e-4, and an exponential learning scheduler with a gamma of 0.998 was applied in all training processes. 
The initial learning rate for training the ResNet and PixBis models on the SynthASpoof database was set to 0.01. The batch size in the training phase was 128 and the training epoch was set to 70. 
Conventional data augmentation techniques: horizontal flipping, scaling and rotating, random gamma adjustment, RGB shifting, and color gittering, were used. The effects of these techniques are explored in Section \ref{ssec:facecrop_aug}. 
In the testing phase, a final PAD decision score of a video is a fused score (mean-rule fusion) of all frames, following \cite{Fang_2022_WACV,DBLP:journals/tbbis/YuLSXZ21,DBLP:conf/icb/GeorgeM19}.

\subsection{Evaluation Metrics}
Following existing cross-domain face PAD methods \cite{Fang_2022_WACV,DBLP:conf/cvpr/LiPWK18,DBLP:conf/cvpr/ShaoLLY19,DBLP:conf/aaai/ShaoLY20}, we report the Half Total Error Rate (HTER), which is the mean of Bona fide Presentation Classification Error Rate (BPCER) \cite{ISO301073} and Attack Presentation Classification Error Rate (APCER) \cite{ISO301073} and Area under the Receiver Operating Characteristic (ROC) Curve (AUC) value for cross-dataset face PAD evaluation. Additionally, ROC curves are illustrated, where the x-axis is APCER and the y-axis is 1-BPCER.

\section{Results} % ready for check
%In this section, we conduct a series of experiments on the SynthASpoof dataset and other widely-used authentic face PAD datasets.
%We first explore the performance gap between the synthetic and the real data by separately training models on synthetic and real datasets, and testing on the real face PAD datasets in Section \ref{ssec:performance_gap}. 
We first explore if the synthetic data can be used to develop PAD solutions. 
We further compare the performance between models trained on the synthetic and the authentic data and tested on different authentic face PAD datasets. 
Then, we investigate the effect of cropping and data augmentation on PAD performance. 
Furthermore, we prove the sanity of adapting MixStyle to enhance the performance of PAD models trained on synthetic data.
%the observable domain gap between synthetic and authentic data is addressed by adapting MixStyle \ref{ssec:effect_mixstyle} to reduce the differences regarding the domain. 
We also explore the usability of SynthASpoof as supplementary training data to enhance the diversity of authentic training data. Finally, a visual analysis and a final discussion is presetned.
 
\subsection{SynthASpoof PAD}
\label{ssec:performance_gap}

%\begin{figure}[th!]
%\begin{center}
%\includegraphics[width=0.99\linewidth]{gap_syn_real.pdf}
%\caption{The performance comparison between models trained on authentic %face PAD datasets (blue) and models trained on SynthASpoof dataset %(green) using the metric of HTER. Models trained on the SynthASpoof %achieved comparable and even better performance in some cases, indicating %the usability of synthetic data for developing face PAD.}
%\label{fig:performance_gap}
%\end{center}
%\end{figure}

\begin{table*}[htb!]
\begin{center}
\resizebox{0.98\textwidth}{!}{
\begin{tabular}{ll|ccc|ccc|ccc|ccc||c}
\hline
Method & Training data              & C → I   & C → M   & C → O  & I → C   & I → M   & I → O  & M → C   & M → I   & M → O   & O → M   & O → C   & O → I   & Average  \\ \hline
\multirow{2}{*}{ResNet} & Authentic  & 38.85 & 18.10 & 17.94 & 42.22 & 18.81 & 28.42 & 27.11 & 16.30 & 30.49 & 15.71 & 23.11 & 23.10  & 25.01 ± 8.74  \\
& SynthASpoof & 8.90  & 25.48 & 34.23 & 39.22 & 25.48 & 34.23 & 39.22 & 8.90  & 34.23 & 25.48 & 39.22 & 8.90   & 26.96 ± 12.04  \\
 \hline
\multirow{2}{*}{PixBis} & Authentic  & 25.05 & 11.19 & 20.72 & 34.22 & 21.67 & 36.57 & 39.11 & 13.65 & 32.58 & 15.00 & 28.11 & 21.80   & 24.97 ± 9.27 
\\
& SynthASpoof & 7.50  & 38.33 & 38.70 & 38.44 & 38.33 & 38.70 & 38.44  & 7.50 & 38.70 & 38.33 & 38.44 & 7.50  & 30.73 ± 14.01
\\ \hline
\end{tabular}}
\end{center}
\caption{The comparison results of models trained on SynthAspoof and authentic datasets, presented as HTER (\%). Models trained on SynthASpoof dataset achieve comparable performances to models trained on authentic datasets in many cases, indicating the usability of SynthASpoof for the development of face PAD.}
\label{tab:comparison_synthetic_authetic}
\end{table*}

To study the feasibility of using SynthASpoof for developing face PAD solutions, we train the models on our SynthASpoof dataset and test on the different authentic datasets from real-world scenarios.  
We also train the models on the authentic data by following the cross-dataset (the evaluation data is unknown) evaluation protocols in recent face PAD works \cite{DBLP:journals/corr/YangLL14,DBLP:journals/tifs/WangHSC21,DBLP:journals/tifs/ZhouLGLLL21}.

In these works \cite{DBLP:journals/corr/YangLL14,DBLP:journals/tifs/WangHSC21,DBLP:journals/tifs/ZhouLGLLL21}, one face PAD dataset is used for the training and the remaining three datasets are separately used as testing data. Therefore, we conduct experiments upon the following 12 scenarios: C → I, C → M, C → O, I → C, I → M, I → O, M → C, M → I, M → O,  O → M, O → I, and O → C. 
Two face PAD models, ResNet-18 and PixBis, are trained following these 12 protocols to evaluate the real-world scenarios. The results are shown in Tab. \ref{tab:comparison_synthetic_authetic}. 
In general, training on SynthASpoof dataset obtains comparable results to the training on authentic data. For example, the average HTER values of ResNet trained on the authentic data and the synthetic data are 25.01\% and 26.96\%, respectively.
When testing on M, C, and O, the models trained on authentic data achieved better performance than the models trained on synthetic data. The possible reason of this observation is that there is a domain gap between the trained synthetic and the tested authentic images. 
%When testing on CASIA dataset, models trained on the synthetic and the authentic datasets achieve both relatively higher HTER values than testing on other test datasets. It might be caused by that CASIA contains a special unknown print attack type, where the eye region is cut out.
When testing on I, models trained on the SynthASpoof obtain significantly better results than models trained on authentic data, because the SynthASpoof dataset includes a diverse range of replay attacks (more than print ones), enabling the model to generalize well on the Idiap ReplayAttack , which mainly consist of replay attacks. This result indicate the applicability of our SynthASpoof dataset for face PAD.

We also visualize the feature distribution using t-SNE \cite{JMLR:v9:vandermaaten08a} on the most challenging case, CASIA, and the best-performing dataset, Idiap ReplayAttack. As shown in Fig.  \ref{fig:tsne} (a) and (c), we have the following observations: 1) Different attack types are clustered separately (represented by different shades of blue), indicating potentially poor generalization on unseen attacks, as evidenced by the results on the CASIA dataset. 2) There is a clear distance between the SynthASpoof and the authentic datasets (represented by different shapes), implying the domain gap between synthetic and authentic data. 

Both quantitative and qualitative results demonstrate the high viability of using SynthASpoof for the development of face PAD algorithms, especially when containing the same type of PA. The observable distance between the synthetic and authentic data will be reduced later in Section \ref{ssec:effect_mixstyle} when incorporating MixStyle.

\subsection{Effect of Cropping and Data Augmentation}
\label{ssec:facecrop_aug}

\begin{table*}[tbh!]
\centering
\resizebox{0.99\textwidth}{!}{
\begin{tabular}{ll|cc|cc|cc|cc|cc}
\hline
\multirow{2}{*}{Margin} & \multirow{2}{*}{Aug} & \multicolumn{2}{c|}{M} & \multicolumn{2}{c|}{C} & \multicolumn{2}{c|}{I} & \multicolumn{2}{c|}{O} & \multicolumn{2}{c}{Average} \\ %\cline{2-11}
& & HTER(\%) $\downarrow$ & AUC(\%) $\uparrow$   & HTER(\%) $\downarrow$ & AUC(\%) $\uparrow$  & HTER(\%) $\downarrow$ & AUC(\%) $\uparrow$    & HTER(\%) $\downarrow$ & AUC(\%) $\uparrow$   & HTER(\%) $\downarrow$ & AUC(\%) $\uparrow$      \\ \hline
0\%  & w/o  & 21.43 &	79.96 &	42.00 &	59.33 &	15.90 &	91.36	& 36.33 &	60.27 &	28.92 &	72.73 \\ \hline
0\%  & w/  & 24.52      & 82.22      & 39.22       & 62.00       & 8.90            & 96.96          & 30.91       & 74.03      & \textbf{25.89}      & \textbf{78.80}   \\ \hline
5\%  & w/  & 24.29      & 79.18      & 47.00       & 52.33       & 11.45           & 94.99          & 35.77       & 67.98      & 29.63      & 73.62   \\ \hline
\end{tabular}}
\caption{The impact of margin extension of face bounding box (extracted from MTCNN) and the effect of using data augmentation by training models on synthetic data and testing separately on four authentic face PAD datasets (M, C, I, and O). The results show that models trained on face images without bounding box extension outperformed models trained on slightly extended face regions. Moreover, applying data augmentation resulted in a better generalized model.}
\label{tab:margin}
\end{table*}

We explore the impact of data augmentation and adding a margin to the face crop by conducting experiments on the synthetic (training) and the authentic data (test). The results are shown in Tab.  \ref{tab:margin}. 
%\textbf{Effect of Augmentation:} 
We use the following data augmentation \footnote{The used augmentation library: Albumentation - \url{https://albumentations.ai/}} techniques: horizontal flipping, scaling and rotation with a limit of 0.1\%, random gamma adjustment within a gamma range from 80 to 180, RGB shifting with a limit of 20, and color jittering with a limit of 0.1\%.
As shown in Tab.  \ref{tab:margin}, using a combined augmentation operation obtains a significant average performance improvement, decreasing the average HTER values from 28.92\% to 25.89\%. 

%\textbf{Effect of Face Crop Ratio:} 
%In \cite{DBLP:conf/cvpr/DamerLFSPB22}, the synthetic face images are cropped by using the MTCCN \cite{DBLP:journals/spl/ZhangZLQ16} bounding box detection with 5\% extension of the width and height. Following this setup, 
As previous work have shown that the consideration area beyond the face is beneficial for PAD performance \cite{DBLP:conf/wacv/NetoSC22}, we argue that this enhancement might be related to properties of specific limited dataset and might not generalize well on unknown data, therefore we study the inclusion of such an area. We compare the results between cropping face region without extension and with 5\% extension of bounding box extracted from MTCNN \cite{DBLP:journals/spl/ZhangZLQ16}. The results in Tab.  \ref{tab:margin} indicate that cropping faces with extension leads to a lower PAD performance on unknown datasets. %One possible reason is that the cluttered background information misleads the training convergence.

As a result, in the following experiments, all models are trained on the cropped faces without bounding box extension and using data augmentation to enhance the PAD generalizability.

\subsection{Effect of MixStyle}
\label{ssec:effect_mixstyle}

\begin{table*}[tbh!]
\centering
\resizebox{0.99\textwidth}{!}{
\begin{tabular}{l|cc|cc|cc|cc|cc}
\hline
\multirow{2}{*}{Method} & \multicolumn{2}{c|}{M} & \multicolumn{2}{c|}{C} & \multicolumn{2}{c|}{I} & \multicolumn{2}{c|}{O} & \multicolumn{2}{c}{Average} \\ %\cline{2-11}
& HTER(\%) $\downarrow$ & AUC(\%) $\uparrow$   & HTER(\%) $\downarrow$ & AUC(\%) $\uparrow$  & HTER(\%) $\downarrow$ & AUC(\%) $\uparrow$    & HTER(\%) $\downarrow$ & AUC(\%) $\uparrow$   & HTER(\%) $\downarrow$ & AUC(\%) $\uparrow$      \\ \hline
ResNet & 25.48 & 79.54 & 39.22 & 62.00	& 8.90 & 96.96	& \textbf{34.23} & \textbf{71.48} & 26.96 & 77.50 \\ 
\quad +MixStyle  & \textbf{21.43} & \textbf{81.97}   & \textbf{22.78}  & \textbf{83.66}      & \textbf{6.70}  & \textbf{98.30}  & 36.07   & 69.52   &   \textbf{21.75}   & \textbf{83.36}   \\ \hline
PixBis & 38.33   & 63.87      & 38.44      & 64.79     & 7.50        & 96.88        & 35.77       & 63.50     & 30.74     & 72.26  \\
\quad +MixStyle & \textbf{30.48}   & \textbf{70.38}     & \textbf{32.00}     & \textbf{69.36}  & \textbf{6.10}          & \textbf{98.29}         & \textbf{34.46} & \textbf{67.71}   & \textbf{25.76}    & \textbf{76.44}  \\ \hline
\end{tabular}}
\caption{The ablation study of MixStyle for adapting models from the synthetic domain to authentic domain. Note that the target domain (authentic face PAD) is used in the training process for MixStyle without labels. The bold number indicates the best performance for each method, pointing out that the usage of MixStyle resulted in an enhanced PAD performance in general.}
\label{tab:mixstyle_effect}
\end{table*}

\begin{figure}[th!]
\begin{center}
\begin{subfigure}[b]{0.8\linewidth}
     \centering
     \includegraphics[width=\linewidth]{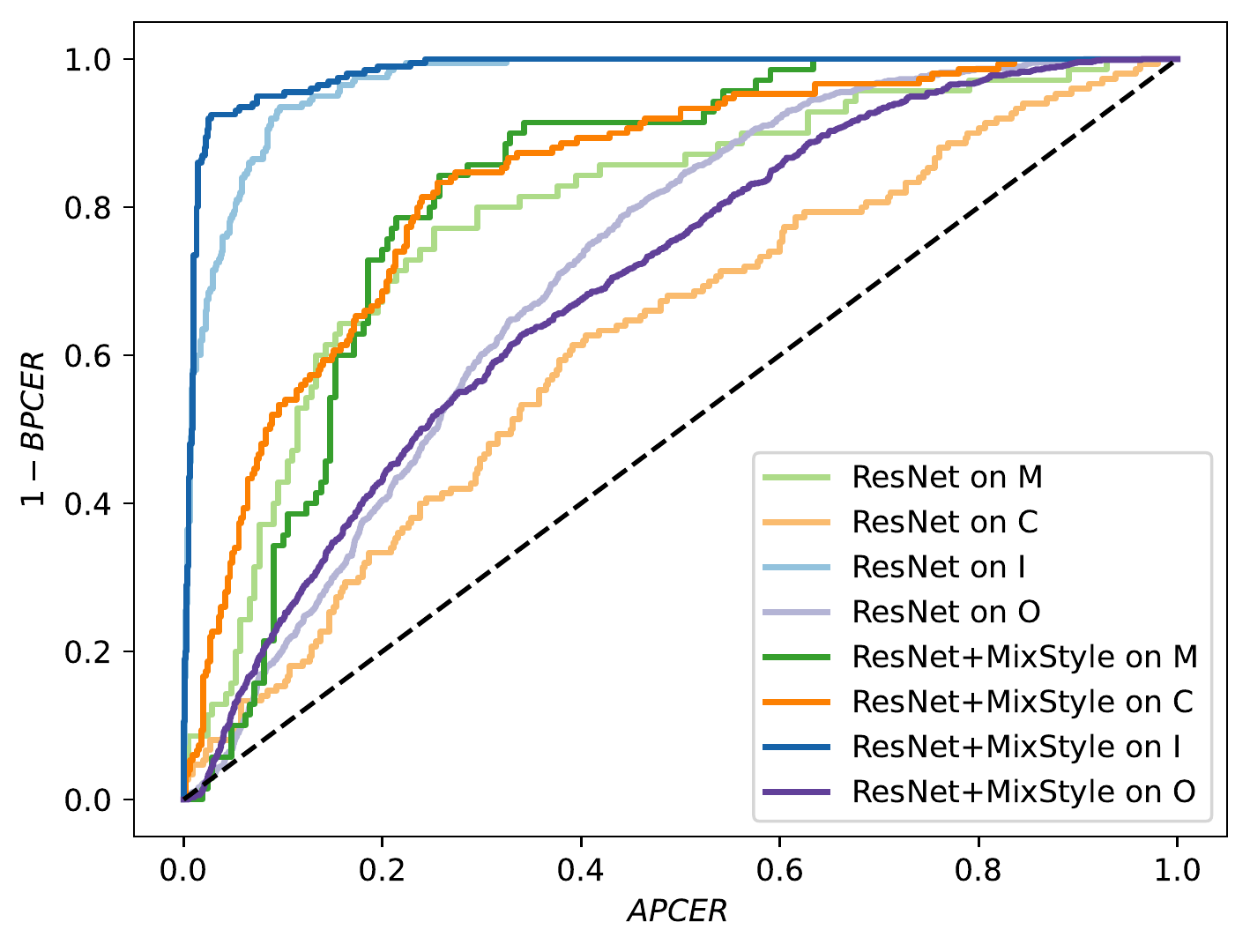}
     \caption{ResNet}
\end{subfigure}

\begin{subfigure}[b]{0.8\linewidth}
     \centering
     \includegraphics[width=\linewidth]{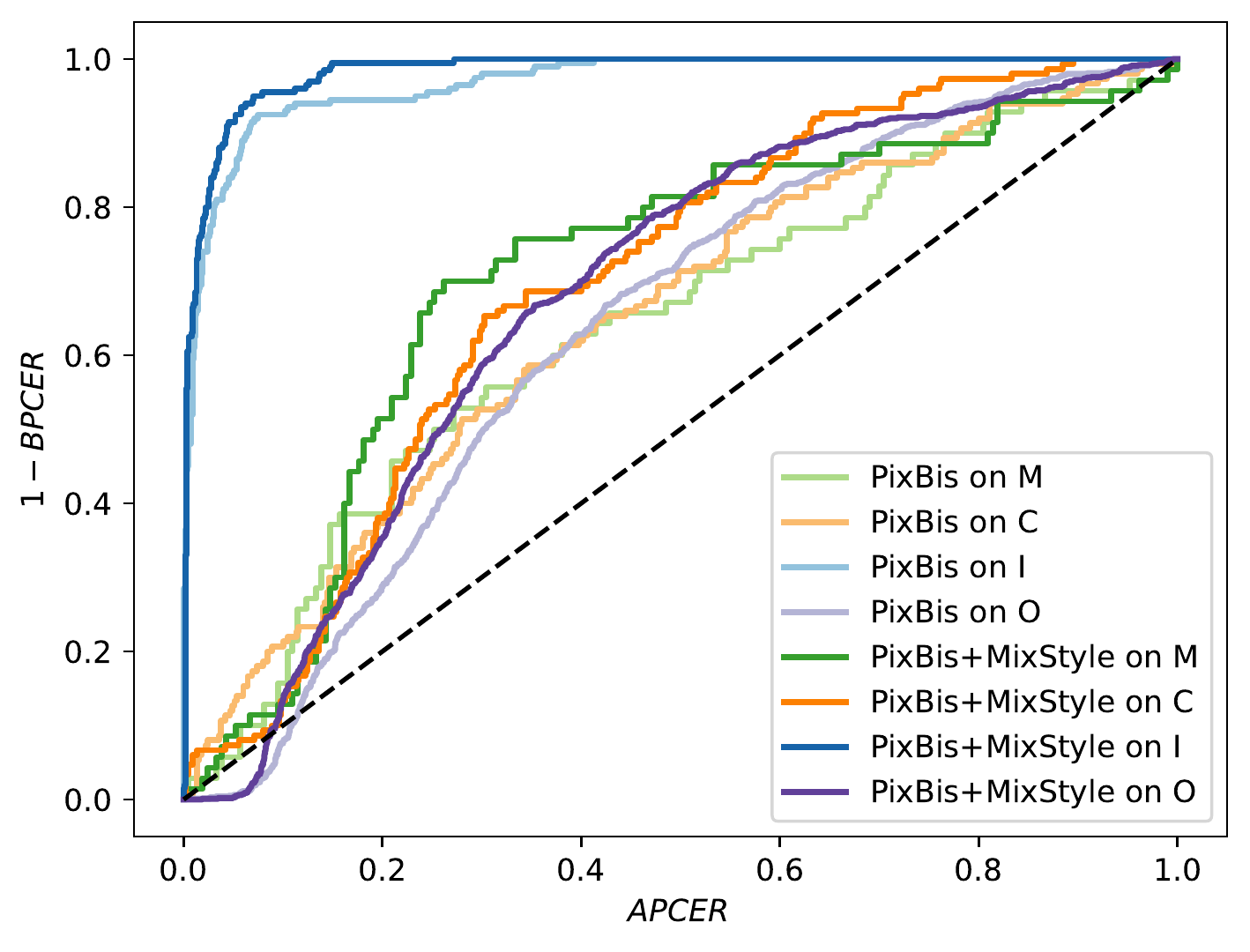}
     \caption{PixBis}

\end{subfigure}

\caption{ROC curves of ResNet (a) and PixBis (b) trained on SynthASpoof dataset and tested on four authentic face PAD datasets (M, C, I, and O). The light colors represent the baseline model and the heavy colors indicate the models trained with the help of MixStyle, mostly leading to better performance (higher curves).}
\label{fig:roc}
\end{center}
\end{figure}

\begin{figure*}[th!]
\begin{center}
\includegraphics[width=0.99\linewidth]{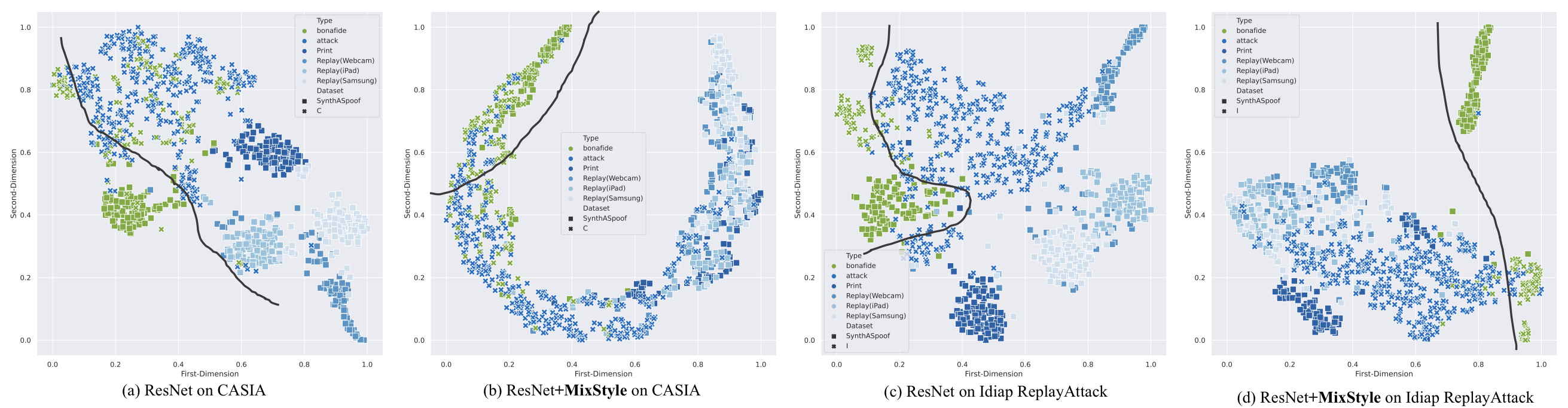}
\caption{Visualization of the feature distribution by using t-SNE \cite{JMLR:v9:vandermaaten08a} for the training on synthetic (\ding{110}) and test on authentic face PAD (\ding{54}) samples, including the most challenging dataset CASIA in our case and the best performing dataset Idiap ReplayAttack. 
The bona fide samples in both datasets are illustrated by green, while different attack types in the SynthASpoof dataset and attacks in the authentic dataset are represented in various shades of blue. Fig.  (a) and (c) demonstrate a clear distance between different attack in the SynthASpoof dataset and a distance between synthetic and authentic data for both classes, bona fides and attack. Fig.  (b) and (c) indicate that MixStyle helps to reduce the distance between the synthetic and the authetic data, i.e., samples within the same class are clustered more closely. } 
\label{fig:tsne}
\vspace{-5mm}
\end{center}
\end{figure*}

As discussed in Sec.  \ref{ssec:performance_gap} and showed in Fig.  \ref{fig:tsne}, there is a distance between the training synthetic and the testing authentic samples.
Motivated by this, MixStyle \cite{DBLP:conf/iclr/ZhouY0X21} is adapted to transfer the domain-related information from the authentic data to the synthetic data. In the training process, the authentic face PAD dataset is used without label only for calculating feature statistics, i.e., the loss is only computed based on the synthetic data.
%The goal of MixStyle here is to reduce the domain gap between the synthetic and the real data by increasing the domain diversity of the SynthASpoof dataset.

Tab.  \ref{tab:mixstyle_effect} shows that applying MixStyle leads to better model generalizability in most cases. The average HTER values on four testing authentic face PAD datasets decreased from 26.96\% to 21.75\% by ResNet and from 30.74\% to 25.76\% by PixBis, while the AUC values increased from 77.50\% to 83.36\% by ResNet and from 72.26\% to 76.44\% by PixBis. The ROC curves shown in Fig.  \ref{fig:roc} illustrate the consistent observation for the baseline model and the model with MixStyle. 

To provide a more detailed understanding of the benefit of MixStyle, we visualize the feature space by ResNet models on our most challenging dataset CAISA and the best performed dataset Idiap ReplayAttack in Fig.  \ref{fig:tsne}. We have the following observations: 1) Features of different types of synthetic attacks (different blue squares) are clustered more closely by applying MixStyle than baseline models, as well as for bona fide, indicating a better generalizability on unknown attack types. 2) Features of authentic data are clustered more closely within the same class of data by applying MixStyle than baseline models. 3) The distance between features of synthetic (\ding{110}) and authentic data (\ding{54}) is visually reduced by using MixStyle.

The quantitative and visual results discussed above demonstrate that MixStyle helps to enhance the PAD performance of models trained on the synthetic data.

%For example, the ResNet-18 model performance on CASIA dataset is improved from 39.22\% to 22.78\% in terms of HTER value and the AUC value is increased from 62.00\% to 83.66\%. 
%Therefore, the distance between some synthetic samples and the real samples is reduced, which benefits the cross-domain classification.

\subsection{Effect of a supplementing Authentic Data with SynthASpoof}
\label{ssec:limited_data}

\begin{table*}[htb!]
\begin{center}
\resizebox{0.98\textwidth}{!}{
\begin{tabular}{l|ccc|ccc|ccc|ccc||c}
\hline
Method               & C → I   & C → M   & C → O  & I → C   & I → M   & I → O  & M → C   & M → I   & M → O   & O → M   & O → C   & O →  I   & Average  \\ \hline
Binary CNN \cite{DBLP:journals/corr/YangLL14}  & 45.80 & 25.60 & 36.40 & 44.40 & 48.60 & 45.40 & 50.10 & 49.90 & 31.40 & 30.20 & 41.20 & 47.40 & 41.37 ± 8.42  \\
ADA \cite{DBLP:conf/icb/WangHSC19}    & 17.50 & \textit{9.30}  & 29.10 & 41.60 & 30.50 & 39.60 & 17.70 & 5.10  & 31.20 & 31.50 & 19.80 & 26.80 & 24.98 ± 11.28 \\
DR-MD-Net \cite{DBLP:conf/cvpr/Wang0SC20} & 26.10 & 20.20 & 24.70 & 39.20 & 23.20 & 33.60 & 34.30 & 8.70  & 31.70 & 22.00 & 21.80 & 27.60 & 26.09 ± 8.05  \\
DR-UDA\cite{DBLP:journals/tifs/WangHSC21}    & \textit{15.60} & \textbf{9.00}  & 28.70 & 34.20 & 29.00 & 38.50 & \textit{16.80} & \textbf{3.00}  & 30.20 & 27.40 & 19.50 & 25.40 & 23.11 ± 10.50 \\
SDFANet \cite{DBLP:journals/tifs/ZhouLGLLL21}  & \textbf{15.50} & 12.14 & \textit{17.08} & 46.11 & 24.29 & 41.56 & \textbf{13.33} & 11.36 & \textbf{18.92} & \textbf{11.67} & 19.33 & 18.71 & 20.83 ± 11.44 \\ \hline
ResNet               & 38.85 & 18.10 & 17.94 & 42.22 & \textit{18.81} & 28.42 & 27.11 & 16.30 & 30.49 & 15.71 & 23.11 & 23.10 & 25.01 ± 8.74  \\
+ SynthASpoof             & 36.80 & 13.10 & 20.88 & 33.33 & \textit{18.81} & 31.65 & 28.67 & 11.85 & 26.53 & 19.05 & 20.44 & 18.95 & 23.34 ± 7.97  \\
+SynthASpoof + MixStyle   & 22.50  & 12.86 & 19.49 & 28.56 & 19.52 & \textit{26.96} & 17.44 & 11.10 & \textit{20.95} & \textit{14.76} & \textbf{18.67} & 22.15 & \textit{19.58} ± \textbf{5.20}  \\  \hline
PixBis               & 25.05 & 11.19 & 20.72 & 34.22 & 21.67 & 36.57 & 39.11 & 13.65 & 32.58 & 15.00 & 28.11 & 21.80 & 24.97 ± 9.27  \\
+ SynthASpoof             & 20.15 & 24.05 & 26.55 & \textit{27.67} & 23.10 & 32.41 & 26.56 & 8.10  & 28.18 & 15.95 & \textit{19.00} & \textit{16.50} & 22.35 ± \textit{6.72}  \\
+SynthASpoof + MixStyle   & 22.02 & 11.19 & \textbf{16.48} & \textbf{23.00} & \textbf{14.05} & \textbf{23.74} & 36.56 & \textit{4.05}  & 28.71 & 15.71 & 21.44 & \textbf{12.95} & \textbf{19.16} ± 8.63 \\ \hline
\end{tabular}}
\end{center}
\vspace{-3mm}
\caption{The results of PAD models trained on a combined training dataset, presented as HTER (\%). Combining SynthASpoof data and the authentic PAD images boost the generalizability of PAD models. Moreover, incorporating MixStyle into the training process leads to even a better generalized PAD models. In comparison to existing works, supplementing the authentic data with SynthASpoof and using MixStyle leads to comparable results and an average performance that goes beyond the latest PAD solutions.
}
\label{tab:db-results}
\end{table*}

As models trained on limited data can easily over-fit the training data and thereby generalize poorly to other domains, we investigate the effect of using the SynthASpoof dataset as a supplementary training data to enhance the diversity of authentic training data.
\textit{+ SynthASpoof} in Tab.  \ref{tab:db-results} refers to that the SynthASpoof is combined with the authentic data in the training process.

As shown in Tab.  \ref{tab:db-results}, including synthetic data in the training process improves the generalizability of the PAD models. For example, the average HTER value of ResNet decreases from 25.01\% to 23.34\%, and of PixBis decreases from 24.97\% to 22.35\%.  These results suggest that adding the SynthASpoof dataset increases the diversity of training samples and thus leads to better representation learning.
Despite the overall improvement, the inclusion of synthetic data did not improve performance in all scenarios. A performance degradation is observed in five out of 12 scenarios when training the ResNet model and in four cases with the PixBis model. This might be caused by the distance between synthetic and authentic face data as we discussed in Sec.  \ref{ssec:performance_gap}. Therefore, we also utilized MixStyle in a combined training process, aiming to narrow the domain gap between the synthetic and authentic data, just as we did in Sec.  \ref{ssec:effect_mixstyle} but with the PAD training here including authentic data.% where the authentic data is used to adapt the model from the synthetic data to the authentic data, we use the MixStyle here to mix the feature statistics of the training synthetic and authentic samples, and hence to improve the generalizability of the trained model.
It can be seen that the model trained with MixStyle generalizes better on unseen test data than the one trained without MixStyle, e.g., the average HTER value of ResNet decreases from 23.34\% to 19.58\% and of PixBis decreases from 22.35\% to 19.16\% with the help of MixStyle. With MixStyle, suplementing the authentic data with SynthASpoof improved the PAD performance in 10 out 12 experimental setups for the ResNet-based PAD.

In summary, incorporating the SynthASpoof dataset seems to diversify the training data, alleviating the overfitting issue caused by limited training data. Furthermore, MixStyle narrows the domain gap between synthetic and authentic data, leading to improved model generalizability.

\subsection{Visualization and Analysis}
We visualized the feature distribution learned by ResNet without MixStyle and with MixStyle in Fig.  \ref{fig:tsne} by considering the most challenging case, SynthASpoof → CASIA, and the best performing case SynthASpoof → Idiap ReplayAttack (both with results presented in Tab. \ref{tab:mixstyle_effect}). To avoid the possible overlapping region and obtain a clear observation, we randomly select 500 samples from each dataset and illustrate their distribution by using t-SNE \cite{JMLR:v9:vandermaaten08a}. Comparing Fig.  \ref{fig:tsne} (a) and (b), and (c) and (d), we found that samples obtained by the model with MixStyle are clustered more closely than baseline models, indicating the effectiveness of MixStyle. Furthermore, applying MixStyle results in a clearer decision boundary given the perspective of the discriminative capability.

\subsection{Discussion}
A extensive experiments successfully demonstrated the feasibility of using the SynthASpoof dataset for the development of face PAD solutions by training as a stand-alone dataset and serving as a supplement to increasing the diversity of limited training data.
Although the goal of this work is not to achieve state-of-the-art PAD performances, a comparison to recent major works using the same experimental protocol is presented in Tab. \ref{tab:db-results}. 
This comparison shows that our PAD trained with the supplement of SynthASpoof and with MixStyle actually outperforms these works in many experimental settings, even leading to a better overall (average) performance.
Due to its privacy-friendly characteristic, the SynthASpoof dataset is made publicly available to the research community. Therefore, SynthASpoof can be extended by collecting more attack data to increase the diversity of attacks (printing, screen and capture devices). 
Moreover, researchers can build PAD datasets to tackle research problems, e.g., the PAD fairness issue  \cite{DBLP:journals/corr/abs-2209-09035} by ensuring a demographically-diverse training data. %fairness assessment which requires datasets with a large diversity of subjects \cite{DBLP:journals/corr/abs-2209-09035}. The other direction is further enhance the generalizability of PAD models trained on the synthetic data. 

\section{Conclusion} % ready for check

To address the ethical and legal challenges associated with the usage, reuse, and sharing of authentic biometric data and motivated by the need for large-scale and diverse PAD development datasets, this work introduced the first privacy-friendly and synthetic-based  dataset, SynthASpoof. The dataset consists of 25,000 bona fide and 78,000 presentation attack samples, which is made publicly available for the research community. We successfully proved the usability of the SynthASpoof dataset for the development of face PADs. 
%Moreover, MixStyle was adapted to address the domain shift between the synthetic and the authentic data, leading to a better PAD generalizability. 
We also showed that SynthASpoof enhanced the generalizability of PAD models by enriching the diversity of the limited authentic data. Furthermore, MixStyle helped to decrease the distance between the synthetic and authentic data, resulting in a more robust and generalized presentation attack detector.

\paragraph{Acknowledgment:}
This research work has been funded by the German Federal Ministry of Education and Research and the Hessen State Ministry for Higher Education, Research and the Arts within their joint support of the National Research Center for Applied Cybersecurity ATHENE.

%%%%%%%%% REFERENCES
{\small
\bibliographystyle{ieee_fullname}
\bibliography{egbib}

\begin{thebibliography}{10}\itemsep=-1pt

\bibitem{oulu_npu}
Zinelabdine Boulkenafet, Jukka Komulainen, Lei Li, Xiaoyi Feng, and Abdenour
  Hadid.
\newblock {OULU-NPU:} {A} mobile face presentation attack database with
  real-world variations.
\newblock In {\em {FG}}, pages 612--618. {IEEE} Computer Society, 2017.

\bibitem{DBLP:conf/icpr/BoutrosDK22}
Fadi Boutros, Naser Damer, and Arjan Kuijper.
\newblock Quantface: Towards lightweight face recognition by synthetic data
  low-bit quantization.
\newblock In {\em 26th International Conference on Pattern Recognition, {ICPR}
  2022, Montreal, QC, Canada, August 21-25, 2022}, pages 855--862. {IEEE},
  2022.

\bibitem{DBLP:journals/corr/abs-2112-06592}
Fadi Boutros, Meiling Fang, Marcel Klemt, Biying Fu, and Naser Damer.
\newblock {CR-FIQA:} face image quality assessment by learning sample relative
  classifiability.
\newblock {\em CoRR}, abs/2112.06592, 2021.

\bibitem{DBLP:conf/icb/BoutrosHSRD22}
Fadi Boutros, Marco Huber, Patrick Siebke, Tim Rieber, and Naser Damer.
\newblock Sface: Privacy-friendly and accurate face recognition using synthetic
  data.
\newblock In {\em {IJCB}}, pages 1--11. {IEEE}, 2022.

\bibitem{DBLP:conf/fgr/BoutrosKFKD23}
Fadi Boutros, Marcel Klemt, Meiling Fang, Arjan Kuijper, and Naser Damer.
\newblock Unsupervised face recognition using unlabeled synthetic data.
\newblock In {\em 17th {IEEE} International Conference on Automatic Face and
  Gesture Recognition, {FG} 2023, Waikoloa Beach, HI, USA, January 5-8, 2023},
  pages 1--8. {IEEE}, 2023.

\bibitem{DBLP:conf/fgr/CaoSXPZ18}
Qiong Cao, Li Shen, Weidi Xie, Omkar~M. Parkhi, and Andrew Zisserman.
\newblock Vggface2: {A} dataset for recognising faces across pose and age.
\newblock In {\em {FG}}, pages 67--74. {IEEE} Computer Society, 2018.

\bibitem{replay_attack}
Ivana Chingovska, Andr{\'{e}} Anjos, and S{\'{e}}bastien Marcel.
\newblock On the effectiveness of local binary patterns in face anti-spoofing.
\newblock In {\em {BIOSIG}}, volume {P-196} of {\em {LNI}}, pages 1--7. {GI},
  2012.

\bibitem{onsyntheticdata_20220922}
{César Augusto Fontanillo López and Abdullah Elbi}.
\newblock On synthetic data: a brief introduction for data protection law
  dummies, 2022.

\bibitem{DBLP:conf/cvpr/DamerLFSPB22}
Naser Damer, C{\'{e}}sar Augusto~Fontanillo L{\'{o}}pez, Meiling Fang,
  No{\'{e}}mie Spiller, Minh~Vu Pham, and Fadi Boutros.
\newblock Privacy-friendly synthetic data for the development of face morphing
  attack detectors.
\newblock In {\em {IEEE/CVF} Conference on Computer Vision and Pattern
  Recognition Workshops, {CVPR} Workshops 2022, New Orleans, LA, USA, June
  19-20, 2022}, pages 1605--1616. {IEEE}, 2022.

\bibitem{DBLP:conf/icb/FangAKD22}
Meiling Fang, Hamza Ali, Arjan Kuijper, and Naser Damer.
\newblock Patchswap: Boosting the generalizability of face presentation attack
  detection by identity-aware patch swapping.
\newblock In {\em {IEEE} International Joint Conference on Biometrics, {IJCB}
  2022, Abu Dhabi, United Arab Emirates, October 10-13, 2022}, pages 1--10.
  {IEEE}, 2022.

\bibitem{DBLP:conf/icb/FangBD22}
Meiling Fang, Fadi Boutros, and Naser Damer.
\newblock Unsupervised face morphing attack detection via self-paced anomaly
  detection.
\newblock In {\em {IEEE} International Joint Conference on Biometrics, {IJCB}
  2022, Abu Dhabi, United Arab Emirates, October 10-13, 2022}, pages 1--11.
  {IEEE}, 2022.

\bibitem{DBLP:conf/fgr/FangBKD21}
Meiling Fang, Fadi Boutros, Arjan Kuijper, and Naser Damer.
\newblock Partial attack supervision and regional weighted inference for masked
  face presentation attack detection.
\newblock In {\em 16th {IEEE} International Conference on Automatic Face and
  Gesture Recognition, {FG} 2021, Jodhpur, India, December 15-18, 2021}, pages
  1--8. {IEEE}, 2021.

\bibitem{Fang_2022_WACV}
Meiling Fang, Naser Damer, Florian Kirchbuchner, and Arjan Kuijper.
\newblock Learnable multi-level frequency decomposition and hierarchical
  attention mechanism for generalized face presentation attack detection.
\newblock In {\em {IEEE/CVF} Winter Conference on Applications of Computer
  Vision, {WACV} 2022, Waikoloa, HI, USA, January 3-8, 2022}, pages 1131--1140.
  {IEEE}, 2022.

\bibitem{DBLP:journals/pr/FangDKK22}
Meiling Fang, Naser Damer, Florian Kirchbuchner, and Arjan Kuijper.
\newblock Real masks and spoof faces: On the masked face presentation attack
  detection.
\newblock {\em Pattern Recognit.}, 123:108398, 2022.

\bibitem{DBLP:journals/corr/abs-2209-09035}
Meiling Fang, Wufei Yang, Arjan Kuijper, Vitomir Struc, and Naser Damer.
\newblock Fairness in face presentation attack detection.
\newblock {\em CoRR}, abs/2209.09035, 2022.

\bibitem{DBLP:conf/icb/GeorgeM19}
Anjith George and S{\'{e}}bastien Marcel.
\newblock Deep pixel-wise binary supervision for face presentation attack
  detection.
\newblock In {\em {ICB}}, pages 1--8. {IEEE}, 2019.

\bibitem{DBLP:journals/tifs/GeorgeMGNAM20}
Anjith George, Zohreh Mostaani, David Geissenbuhler, Olegs Nikisins,
  Andr{\'{e}} Anjos, and S{\'{e}}bastien Marcel.
\newblock Biometric face presentation attack detection with multi-channel
  convolutional neural network.
\newblock {\em {IEEE} Trans. Inf. Forensics Secur.}, 15:42--55, 2020.

\bibitem{DBLP:conf/eccv/GuoZHHG16}
Yandong Guo, Lei Zhang, Yuxiao Hu, Xiaodong He, and Jianfeng Gao.
\newblock Ms-celeb-1m: {A} dataset and benchmark for large-scale face
  recognition.
\newblock In {\em {ECCV} {(3)}}, volume 9907 of {\em Lecture Notes in Computer
  Science}, pages 87--102. Springer, 2016.

\bibitem{DBLP:conf/cvpr/HeZRS16}
Kaiming He, Xiangyu Zhang, Shaoqing Ren, and Jian Sun.
\newblock Deep residual learning for image recognition.
\newblock In {\em {CVPR}}, pages 770--778. {IEEE} Computer Society, 2016.

\bibitem{DBLP:conf/cvpr/HuangLMW17}
Gao Huang, Zhuang Liu, Laurens van~der Maaten, and Kilian~Q. Weinberger.
\newblock Densely connected convolutional networks.
\newblock In {\em 2017 {IEEE} Conference on Computer Vision and Pattern
  Recognition, {CVPR} 2017, Honolulu, HI, USA, July 21-26, 2017}, pages
  2261--2269. {IEEE} Computer Society, 2017.

\bibitem{DBLP:conf/icb/HuberBLRRDNGSCT22}
Marco Huber, Fadi Boutros, Anh~Thi Luu, Kiran~B. Raja, Raghavendra Ramachandra,
  Naser Damer, Pedro~C. Neto, Tiago Gon{\c{c}}alves, Ana~F. Sequeira, Jaime~S.
  Cardoso, Jo{\~{a}}o Tremo{\c{c}}o, Miguel Louren{\c{c}}o, Sergio Serra,
  Eduardo Cerme{\~{n}}o, Marija Ivanovska, Borut Batagelj, Andrej Kronovsek,
  Peter Peer, and Vitomir Struc.
\newblock {SYN-MAD} 2022: Competition on face morphing attack detection based
  on privacy-aware synthetic training data.
\newblock In {\em {IJCB}}, pages 1--10. {IEEE}, 2022.

\bibitem{ISO301073}
{International Organization for Standardization}.
\newblock {ISO/IEC DIS 30107-3:2016: Information Technology – Biometric
  presentation attack detection – P. 3: Testing and reporting}, 2017.

\bibitem{DBLP:conf/nips/KarrasAHLLA20}
Tero Karras, Miika Aittala, Janne Hellsten, Samuli Laine, Jaakko Lehtinen, and
  Timo Aila.
\newblock Training generative adversarial networks with limited data.
\newblock In {\em NeurIPS}, 2020.

\bibitem{DBLP:conf/cvpr/KarrasLA19}
Tero Karras, Samuli Laine, and Timo Aila.
\newblock A style-based generator architecture for generative adversarial
  networks.
\newblock In {\em {CVPR}}, pages 4401--4410. Computer Vision Foundation /
  {IEEE}, 2019.

\bibitem{DBLP:conf/cvpr/Kemelmacher-Shlizerman16}
Ira Kemelmacher{-}Shlizerman, Steven~M. Seitz, Daniel Miller, and Evan
  Brossard.
\newblock The megaface benchmark: 1 million faces for recognition at scale.
\newblock In {\em {CVPR}}, pages 4873--4882. {IEEE} Computer Society, 2016.

\bibitem{DBLP:conf/cvpr/LiPWK18}
Haoliang Li, Sinno~Jialin Pan, Shiqi Wang, and Alex~C. Kot.
\newblock Domain generalization with adversarial feature learning.
\newblock In {\em {CVPR}}, pages 5400--5409. {IEEE} Computer Society, 2018.

\bibitem{DBLP:journals/tifs/LiuZYWSLTEXLGLL22}
Ajian Liu, Chenxu Zhao, Zitong Yu, Jun Wan, Anyang Su, Xing Liu, Zichang Tan,
  Sergio Escalera, Junliang Xing, Yanyan Liang, Guodong Guo, Zhen Lei, Stan~Z.
  Li, and Du Zhang.
\newblock Contrastive context-aware learning for 3d high-fidelity mask face
  presentation attack detection.
\newblock {\em {IEEE} Trans. Inf. Forensics Secur.}, 17:2497--2507, 2022.

\bibitem{DBLP:conf/cvpr/LiuSJ019}
Yaojie Liu, Joel Stehouwer, Amin Jourabloo, and Xiaoming Liu.
\newblock Deep tree learning for zero-shot face anti-spoofing.
\newblock In {\em {IEEE} Conference on Computer Vision and Pattern Recognition,
  {CVPR} 2019, Long Beach, CA, USA, June 16-20, 2019}, pages 4680--4689.
  Computer Vision Foundation / {IEEE}, 2019.

\bibitem{DBLP:conf/wacv/NetoSC22}
Pedro~C. Neto, Ana~F. Sequeira, and Jaime~S. Cardoso.
\newblock Myope models - are face presentation attack detection models
  short-sighted?
\newblock In {\em {IEEE/CVF} Winter Conference on Applications of Computer
  Vision Workshops, {WACV} - Workshops, Waikoloa, HI, USA, January 4-8, 2022},
  pages 390--399. {IEEE}, 2022.

\bibitem{DBLP:conf/iccv/QiuYG00T21}
Haibo Qiu, Baosheng Yu, Dihong Gong, Zhifeng Li, Wei Liu, and Dacheng Tao.
\newblock Synface: Face recognition with synthetic data.
\newblock In {\em {ICCV}}, pages 10860--10870. {IEEE}, 2021.

\bibitem{padisi-Face}
Mohammad Rostami, Leonidas Spinoulas, Mohamed~E. Hussein, Joe Mathai, and Wael
  Abd{-}Almageed.
\newblock Detection and continual learning of novel face presentation attacks.
\newblock In {\em {ICCV}}, pages 14831--14840. {IEEE}, 2021.

\bibitem{DBLP:conf/cvpr/ShaoLLY19}
Rui Shao, Xiangyuan Lan, Jiawei Li, and Pong~C. Yuen.
\newblock Multi-adversarial discriminative deep domain generalization for face
  presentation attack detection.
\newblock In {\em {CVPR}}, pages 10023--10031. Computer Vision Foundation /
  {IEEE}, 2019.

\bibitem{DBLP:conf/aaai/ShaoLY20}
Rui Shao, Xiangyuan Lan, and Pong~C. Yuen.
\newblock Regularized fine-grained meta face anti-spoofing.
\newblock In {\em {AAAI}}, pages 11974--11981. {AAAI} Press, 2020.

\bibitem{JMLR:v9:vandermaaten08a}
Laurens van~der Maaten and Geoffrey Hinton.
\newblock Visualizing data using t-sne.
\newblock {\em Journal of Machine Learning Research}, 9(86):2579--2605, 2008.

\bibitem{DBLP:conf/icb/WangHSC19}
Guoqing Wang, Hu Han, Shiguang Shan, and Xilin Chen.
\newblock Improving cross-database face presentation attack detection via
  adversarial domain adaptation.
\newblock In {\em {ICB}}, pages 1--8. {IEEE}, 2019.

\bibitem{DBLP:conf/cvpr/Wang0SC20}
Guoqing Wang, Hu Han, Shiguang Shan, and Xilin Chen.
\newblock Cross-domain face presentation attack detection via multi-domain
  disentangled representation learning.
\newblock In {\em {CVPR}}, pages 6677--6686. Computer Vision Foundation /
  {IEEE}, 2020.

\bibitem{DBLP:journals/tifs/WangHSC21}
Guoqing Wang, Hu Han, Shiguang Shan, and Xilin Chen.
\newblock Unsupervised adversarial domain adaptation for cross-domain face
  presentation attack detection.
\newblock {\em {IEEE} Trans. Inf. Forensics Secur.}, 16:56--69, 2021.

\bibitem{msu_mfs}
Di Wen, Hu Han, and Anil~K. Jain.
\newblock Face spoof detection with image distortion analysis.
\newblock {\em {IEEE} Trans. Inf. Forensics Secur.}, 10(4):746--761, 2015.

\bibitem{DBLP:journals/corr/YangLL14}
Jianwei Yang, Zhen Lei, and Stan~Z. Li.
\newblock Learn convolutional neural network for face anti-spoofing.
\newblock {\em CoRR}, abs/1408.5601, 2014.

\bibitem{DBLP:journals/tbbis/YuLSXZ21}
Zitong Yu, Xiaobai Li, Jingang Shi, Zhaoqiang Xia, and Guoying Zhao.
\newblock Revisiting pixel-wise supervision for face anti-spoofing.
\newblock {\em {IEEE} Trans. Biom. Behav. Identity Sci.}, 3(3):285--295, 2021.

\bibitem{DBLP:journals/spl/ZhangZLQ16}
Kaipeng Zhang, Zhanpeng Zhang, Zhifeng Li, and Yu Qiao.
\newblock Joint face detection and alignment using multitask cascaded
  convolutional networks.
\newblock {\em {IEEE} Signal Proc. Lett.}, 23(10):1499--1503, 2016.

\bibitem{DBLP:conf/cvpr/ZhangWLZ0ESWL19}
Shifeng Zhang, Xiaobo Wang, Ajian Liu, Chenxu Zhao, Jun Wan, Sergio Escalera,
  Hailin Shi, Zezheng Wang, and Stan~Z. Li.
\newblock A dataset and benchmark for large-scale multi-modal face
  anti-spoofing.
\newblock In {\em {CVPR}}, pages 919--928. Computer Vision Foundation / {IEEE},
  2019.

\bibitem{DBLP:conf/eccv/ZhangYLYYSL20}
Yuanhan Zhang, Zhenfei Yin, Yidong Li, Guojun Yin, Junjie Yan, Jing Shao, and
  Ziwei Liu.
\newblock Celeba-spoof: Large-scale face anti-spoofing dataset with rich
  annotations.
\newblock In {\em Computer Vision - {ECCV} 2020 - 16th European Conference,
  Glasgow, UK, August 23-28, 2020, Proceedings, Part {XII}}, volume 12357 of
  {\em Lecture Notes in Computer Science}, pages 70--85. Springer, 2020.

\bibitem{casia_fas}
Zhiwei Zhang, Junjie Yan, Sifei Liu, Zhen Lei, Dong Yi, and Stan~Z. Li.
\newblock A face antispoofing database with diverse attacks.
\newblock In {\em {ICB}}, pages 26--31. {IEEE}, 2012.

\bibitem{DBLP:conf/iclr/ZhouY0X21}
Kaiyang Zhou, Yongxin Yang, Yu Qiao, and Tao Xiang.
\newblock Domain generalization with mixstyle.
\newblock In {\em {ICLR}}. OpenReview.net, 2021.

\bibitem{DBLP:journals/tifs/ZhouLGLLL21}
Lifang Zhou, Jun Luo, Xinbo Gao, Weisheng Li, Bangjun Lei, and Jiaxu Leng.
\newblock Selective domain-invariant feature alignment network for face
  anti-spoofing.
\newblock {\em {IEEE} Trans. Inf. Forensics Secur.}, 16:5352--5365, 2021.

\end{thebibliography}
}

\end{document}